\documentclass{article}
\usepackage{arxiv}

\usepackage{tabularx}
\usepackage{dirtytalk}
\usepackage{quoting}
\usepackage{subcaption}
\usepackage{amsmath}
\usepackage{url}
\usepackage{hyperref}
\usepackage{cleveref}
\usepackage{natbib}

\quotingsetup{font=itshape,vskip=1pt, leftmargin=15pt}
\usepackage[most]{tcolorbox}

\newtcolorbox{mybox}{
enhanced,
boxrule=0pt,frame hidden,
borderline west={4pt}{0pt}{white!75!black},
colback=white,
sharp corners
}
\AtBeginDocument{%
  }

\begin{document}

\newcommand{\samplemean}{\bar{x}}
\newcommand{\samplesd}{s}
\newcommand{\chatbot}{\emph{WAgent}}
\newcommand{\website}{\emph{Website}}
\newcommand{\studyOne}{\emph{Study 1}}
\newcommand{\studyTwo}{\emph{Study 2}}

\newcommand{\todol}[1]{\textsf{\textbf{\textcolor{green!55!blue}{[Lars: #1]}}}}
\newcommand{\todoj}[1]{\textsf{\textbf{\textcolor{yellow!55!red}{[Jakob: #1]}}}}
\newcommand{\todod}[1]{\textsf{\textbf{\textcolor{yellow!55!green}{[Daniel: #1]}}}}

\title{Towards Sustainable Web Agents: A Plea for Transparency and Dedicated Metrics for Energy Consumption}
\renewcommand{\shorttitle}{Towards Sustainable Web Agents}

\author{ \href{https://orcid.org/0000-0001-6294-2915}{\includegraphics[scale=0.06]{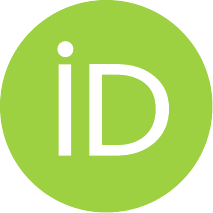}\hspace{1mm}Lars Krupp}\\
	German Research Center for Artificial Intelligence (DFKI),\\
    RPTU Kaiserslautern-Landau\\
	\texttt{lars.krupp@dfki.de} \\
	%% examples of more authors
	\And
	\href{https://orcid.org/0000-0003-2643-4504}{\includegraphics[scale=0.06]{orcid.pdf}\hspace{1mm}Daniel Geißler} \\
	German Research Center for Artificial Intelligence (DFKI)\\
    \And
	\href{https://orcid.org/0000-0003-0320-6656}{\includegraphics[scale=0.06]{orcid.pdf}\hspace{1mm}Paul Lukowicz} \\
	German Research Center for Artificial Intelligence (DFKI),\\
    RPTU Kaiserslautern-Landau\\
    \And
	\href{https://orcid.org/0000-0002-0698-7470}{\includegraphics[scale=0.06]{orcid.pdf}\hspace{1mm}Jakob Karolus} \\
	German Research Center for Artificial Intelligence (DFKI),\\
    RPTU Kaiserslautern-Landau\\
	%\texttt{stariate@ee.mount-sheikh.edu} \\
	%% \AND
	%% Coauthor \\
	%% Affiliation \\
	%% Address \\
	%% \texttt{email} \\
	%% \And
	%% Coauthor \\
	%% Affiliation \\
	%% Address \\
	%% \texttt{email} \\
	%% \And
	%% Coauthor \\
	%% Affiliation \\
	%% Address \\
	%% \texttt{email} \\
}

% <concept_id>10003120.10003121</concept_id>
% <concept_desc>Human-centered computing~Human computer interaction (HCI)</concept_desc>

 \maketitle

% \ccsdesc[500]{Human-centered computing~Human computer interaction (HCI)}


\begin{abstract}

Improvements in the area of large language models have shifted towards the construction of models capable of using external tools and interpreting their outputs. These so-called web agents have the ability to interact autonomously with the internet.
This allows them to become powerful daily assistants handling time-consuming, repetitive tasks while supporting users in their daily activities. 
While web agent research is thriving, the sustainability aspect of this research direction remains largely unexplored.  We provide an initial exploration of the energy and $CO_2$ cost associated with web agents. Our results show how different philosophies in web agent creation can severely impact the associated expended energy. We highlight lacking transparency regarding the disclosure of model parameters and processes used for some web agents as a limiting factor when estimating energy consumption. As such, our work advocates a change in thinking when evaluating web agents, warranting dedicated metrics for energy consumption and sustainability.

%We shine a light on the difficulties associated with obtaining all relevant information to make reasonable estimations of these quantities and show that the sustainability aspect of web agents should be a metric tested against.

\end{abstract}



\section{Introduction}
Web agents are LLM-powered systems capable of "browsing the web". Considered the next milestone in advancing large language models~\cite{projectmariner,openaiagent}, the idea enables automated interaction with the internet. In essence, allowing web agents to explore much like a human would. %This grounds them with up-to-date information, grounding their responses. %This automated interaction also encompasses tasks like shopping, where the agents are capable to find, compare, select, and buy different articles depending on a user query. % To do this, they press buttons, use search bars and navigate menus.
While this concept has offered vast potential for intelligent tools, high computational costs~\cite{Samsi2023From} still remain a major issue. Context window sizes have exploded, exponentially increasing energy consumption and questioning the sustainability of this research direction.

LASER~\cite{ma2023laser} is a popular web agent, able to achieve a success rate  on the WebShop~\cite{yao2022webshop} benchmark equivalent to the average success rate achieved by humans. Its capabilities include the ability to automatically find, select and buy a product on a website depending on a user query. The agent is capable to find, compare, select, and buy different articles. To do this, the agent presses buttons, uses search bars and navigates menus. While this autonomous interaction is impressive, so is its energy bill. An estimated 2930Wh per action (button press, using a search bar, ...) on the Mind2Web~\cite{deng2024mind2web} benchmark. With this, using LASER~\cite{ma2023laser} emits approximately 9.6\,kg $CO_2$ per task on the Mind2Web benchmark. This is equivalent to a car driving 39\,km (nearly the distance of a marathon). We argue that such approaches are not sustainable in the long run and advocate for a more holistic evaluation of web agent and in particular LLM performances, including energy spent to achieve benchmark results.

In this work, we present an illustrative comparison between two popular web agents. MindAct~\cite{deng2024mind2web}, a web agent using open source LLMs and smart preprocessing to reduce its energy bill and the aforementioned LASER~\cite{ma2023laser}. By estimating and comparing the amount of energy consumed by these web agents with vastly different design philosophies, we highlight the impact of a web agent's design on its energy consumption. Using a conservative estimation (\Cref{sec:mindact}), LASER spends approximately 1500 times the energy than MindAct. Further, we point out the general difficulties with energy consumption estimation due to a lack of model parameter disclosure and propose which values should be reported in the future when releasing new web agents to make it possible to compare web agents based on their environmental impact.

Our work encourages a change in thinking on how we evaluate web agent performance and disclosure of model parameters, reprimanding missing transparency. We show the stark differences in energy consumption between state-of-the-art web agents and propose metrics to report which allow a straightforward comparison of their energy efficiency. We advocate for a holistic evaluation of web agents incorporating dedicated metrics for energy consumption.

\section{Related Work}
With the rapid improvements of LLMs in recent years~\cite{dubey2024llama} and their ever improving capabilities in tool-use~\cite{dubey2024llama} a new frontier of research has become possible. With the goal of building agents that can interact with the internet much like a human would, web agents recently are gaining traction~\cite{deng2024mind2web, yao2022webshop}. In this newly developing field, many different benchmarks are being proposed in rapid succession~\cite{yao2022webshop,deng2024mind2web,he2024webvoyagerbuildingendtoendweb}. While efforts to unify these benchmarks exist~\cite{dechezelles2024browsergymecosystemwebagent}, at the moment, comparing the performance of different web agents is often not feasible. 
Analogously, the approaches in web agent construction show a high diversity. While some only use HTML for their input~\cite{ma2023laser, deng2024mind2web}, others use the accessibility tree instead~\cite{dechezelles2024browsergymecosystemwebagent} or supplement it with screenshots~\cite{zheng2024gpt} or even use screenshots exclusively, like Pix2Act~\cite{lu2024weblinx}. This extends into the approaches used for preprocessing~\cite{deng2024mind2web,gur2023real}, the integration of memory modules~\cite{ma2023laser} and which kind of language models are used. While some approaches use open source models~\cite{deng2024mind2web,gur2023real} many use proprietary models~\cite{ma2023laser, zheng2024gpt,yang2024agentoccamsimplestrongbaseline,zhang2024webpilotversatileautonomousmultiagent} making a precise evaluation of the environmental impact difficult.

\subsection{Environmental Impact of LLMs}
\label{sec:enviroment}
There is a pressing discussion on sustainability and the environmental impact of developing and deploying LLMs~\cite{bender2021dangers}.
Since LLMs are trained on massive amounts of data to provide thorough knowledge, large-scale data centers are required to train the complex model architectures in sufficient time properly.
Even though detailed information about LLMs are usually not publicly available, data from previous versions such as OpenAIs GPT-3 already state the usage of 175 billion model parameters being trained on 570 GB of data~\cite{brown2020language}.

Statistics on the energy consumption are even sparser and commonly rely on rough estimations due to multiple unknown factors, such as the hardware architecture and efficiency, the training and optimization strategies, and most importantly the energy mix.
In work by ~\citet{lannelongue2021green}, a Green Algorithm Calculator is proposed to estimate the Carbon Footprint of LLM training.
Depending on the utilized hardware's location, the energy mix between fossil and renewable energies severely impacts the environmental stress. For instance, 20g CO2e kWh (carbon dioxide
equivalent per kilowatt hour) in Norway and Switzerland to over 800g CO2e kWh in Australia, South Africa, and the USA.
Based on the BERT model~\cite{devlin2018bert}, trained in the USA, they calculated a potential environmental impact of 0.754 metric tons of CO2 for a single training of 79h on 64 Tesla V100 GPUs with an average utilization of 62.7\%.
For the popular GPT-3 model, also based on the USA energy mix, it is estimated that around 550 metric tons of CO2 emissions were produced to complete the full training, exceeding the previous estimations from BERT tremendously due to the complexity and dataset size increase~\cite{shi2023thinking}.
On top of the single training run, usually, a significant amount of energy is wasted on ineffective versions of the LLM or for tuning the hyperparameter spaces~\cite{Verdecchia2023A}.

After training, the energy demand for deploying LLMs to the public additionally contributes to the overall environmental stress.
According to \citet{Samsi2023From}, the energy demand for the inference depends on the utilization of the hardware setup since requests should be properly scheduled to utilize the GPUs in their most efficient window.
They propose the energy per token as a metric to compare performance with sustainability, especially for comparing quantized LLM versions, since the overall energy demand for inference depends on various unpredictable factors such as the number of users and the duration of the deployment phase.
Additionally, the environmental impact extends beyond energy consumption to include resource use, such as the water needed for cooling data centers, and the e-waste generated by the disposal of outdated hardware, highlighting the complexity and difficulty of calculating and comparing the LLM carbon footprint throughout the whole life cycle~\cite{patterson2021carbon}.
Within our work, we aim to quantify the energy consumption, carbon footprint and offer an insight on the enhanced sustainability potential of web agents.

\section{Estimating the Energy Consumption of Web Agents}
\begin{figure}[h]
    \centering
    \includegraphics[width=0.5\linewidth]{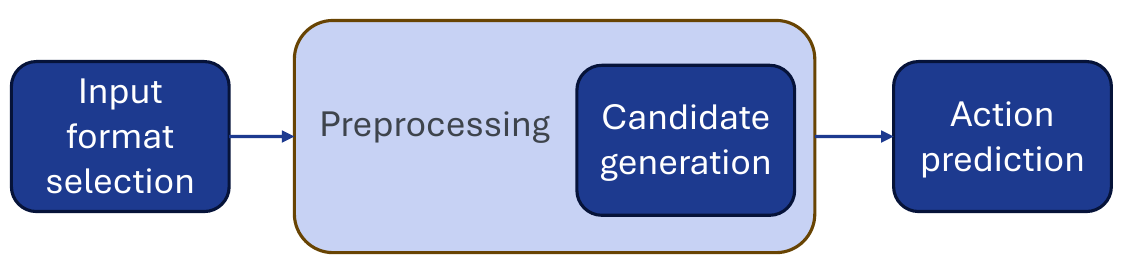}
    \caption{A pipeline depicting the generic structure of web agents.}
    \label{fig:overview_pipe}
\end{figure}

Web agents are capable of autonomously navigating websites based on a task given by the user. To do this, they execute actions, like using a search field or pressing a button. Which action to take to advance is a complex challenge that current approaches try to solve using LLMs. To decrease this complexity, many approaches~\cite{deng2024mind2web,gur2023real} employ various preprocessing steps to generate likely candidates for actions. From these candidates the action prediction model then selects the action to take. \Cref{fig:overview_pipe} represents a generic pipeline for web agents.

\label{sec:evaluation}

With MindAct~\cite{deng2024mind2web} and LASER~\cite{ma2023laser} we chose two web agents differing in many aspects to provide an example of the diversity in approaches and mentalities present in the field. While MindAct uses comparatively small open-source models and does extensive preprocessing to get the best possible performance out of the available resources, LASER uses a proprietary model at its core with minimal preprocessing being done. The difference in philosophy also shows in the resulting energy cost.

To estimate the energy cost for each model, we analyzed and dissected available resources such as the accompanying publications~\cite{deng2024mind2web,ma2023laser} and the publicly available source code (\url{https://github.com/OSU-NLP-Group/Mind2Web}, \url{https://github.com/Mayer123/LASER}). For the open-source models (used in MindAct), we set up a custom test environment to measure their energy consumption.

For our evaluation, we decided to use the popular Mind2Web~\cite{deng2024mind2web} benchmark. In contrast to other web agent benchmarks~\cite{yao2022webshop,liu2018reinforcement}, Mind2Web consists of real-world websites. This allows for a realistic evaluation with respect to the average number of tokens per website. Mind2Web consists of 2350 tasks on 137 websites in 31 domains with an average number of actions needed for task completion of 7.3 and an average number of 1135 HTML elements per website. 

%Since different language models use different embedding models, the number of tokens for the same text differs from model to model. We calculated the average number of tokens contained within a HTML page for the Mind2Web~\cite{deng2024mind2web} benchmark to be $T_{MW_{DeBERTa}}=118798$ tokens using the DeBERTa tokenizer~\cite{he2021debertav3} and $t_{MW_GPT-4} = 93778$ tokens using the tokenizer of GPT-4~\cite{openai2024gpt4technicalreport}.

\subsection{MindAct}
\label{sec:mindact}
\begin{figure}[ht]
    \centering
    \includegraphics[width=1\linewidth]{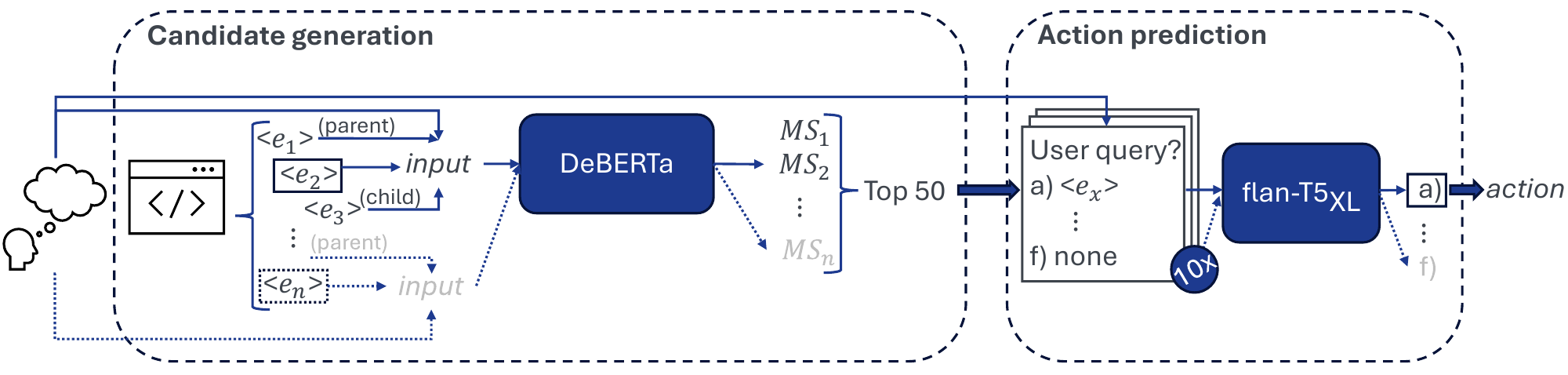}
    \caption{Pipeline depicting how an action is chosen in MindAct following the two stage process shown in \Cref{fig:overview_pipe}. For each element $e_i$ in the DOM, the user query and textual descriptions of parent and child elements are added to form the \textit{input} of DeBERTa which calculates a matching score $MS_i$. The 50 highest scored elements are selected for the next stage and transformed into 10 multiple-choice question answering tasks with the user query as the question and given to flan-T5$_{XL}$ which selects an \textit{action} to take.}
    \label{fig:mindact_pipe}
\end{figure}

MindAct~\cite{deng2024mind2web} divides the process of finding the correct \textit{action} to advance with fulfilling a given task on the web into two stages, as depicted in \Cref{fig:mindact_pipe}. The first stage, called candidate generation, is treated as a ranking task. Here, their finetuned DeBERTa 86M~\cite{he2021debertav3} model is given as \textit{input}: the user query, previous actions and a cleaned representation of each element in the Document Object Model (DOM) of the HTML webpage. Each cleaned element consists of its tag, its textual content, its salient attribute values and a textual representation of its respective parent and child nodes. From this, DeBERTa estimates a matching value $MS_i$ between 0 and 1 --- indicating how well an element matches to the given user query. This process is repeated for all elements in the DOM. We estimate that the total number of tokens processed by DeBERTa at the end of this process is at the very minimum equivalent to the amount of tokens in the original HTML (no parent/child elements for any DOM element) and at most three times the original HTML tokens (all DOM elements have both parent and child elements). After processing all elements, the 50 elements with the highest matching values are used in the second stage. 
The second stage, called action prediction, is constructed as a multiple-choice question answering challenge. A finetuned flan-T5$_{XL}$ model is given the user query, five of the returned elements and a none element and is tasked to decide which element is most likely to help towards answering the user query (the \textit{action} to perform). This is done a minimum of 10 times until all 50 returned elements are processed. If more than one not-none answer was selected, those elements are processed again until only one element remains or every element was rejected. For our estimation, we assume the maximum input length of flan-T5$_{XL}$~\cite{raffel2020exploring} (512 tokens) for one  multiple-choice question and that a final result is obtained after the first pass (querying flan-T5$_{XL}$ 10 times).

To estimate the amount of energy used to produce one action for the Mind2Web dataset, we first need to know the expended energy per token for the DeBERTa and flan-T5$_{XL}$ models. This expended energy depends on the number of tokens processed by an LLM which changes depending on the tokenizer used by the specific model. We calculated the average number of tokens contained within a HTML page for the Mind2Web~\cite{deng2024mind2web} benchmark to be $\overline{N}_{DeBERTa}=118798$ tokens using the DeBERTa tokenizer~\cite{he2021debertav3}. Using our estimations for DeBERTa ($1 \le k \le 3$) and flan-T5$_{XL}$ (first pass, maximum input length of 512 tokens), this results in the following calculation for the energy cost $E_{action}$ for predicting a single action using MindAct.

\begin{align}
E_{action} &= E_{first\ stage} + 10 \cdot E_{second\ stage}\\
E_{action} &= (k \cdot \overline{N}_{DeBERTa} \cdot e_{DeBERTa}) + (10 \cdot \overline{N}_{flan-T5_{XL}} \cdot e_{flan-T5_{XL}})\\
E_{action} &= (k \cdot 118798 \cdot e_{DeBERTa}) + (10 \cdot 512 \cdot e_{flan-T5_{XL}}) \label{firstline}\\
\intertext{To acquire the energy per token for both models, $e_{DeBERTa}$ and $e_{flan-T5_{XL}}$, we set up an experimental setup using carbontracker~\cite{anthony2020carbontracker}. We conducted a number of runs with a fixed token size for each model on a NVIDIA GeForce RTX 3080Ti. Given the number of runs and the number of tokens per run, we calculated the energy spent per token for the two models, resulting in: }
%e_{DeBERTa} &= \frac{e_{avg}}{n_{tokens}} =\frac{e_{total}}{n_{total-runs}}:n_{token}\\
%&= \frac{0.00188kWh}{1000E}:470token\\
%&=\frac{0.00188Wh}{470token}\\
%e_{DeBERTa}&=0.000004\frac{Wh}{token}\\
e_{DeBERTa}&=4\cdot10^{-6}\,Wh
\intertext{and: }
e_{flan-T5_{XL}}&=102\cdot 10^{-6}\,Wh\\
%e_{flan-T5_{XL}} &=\frac{e_{avg}}{n_{tokens}} =\frac{e_{total}}{n_{total-runs}}:n_{token}\\
%&= \frac{0.00498kWh}{100E} :490token\\
%&=\frac{0.0498Wh}{490token}\\
%&=\underline{\underline{0.000102}}\frac{Wh}{token}\\
\intertext{Entering those into \ref{firstline}:}
E_{action} &= (k \cdot 118798 \cdot 4\cdot 10^{-6}\,Wh) + (10 \cdot 512 \cdot 102\cdot 10^{-6}\,Wh)\\
E_{action} &= (k \cdot 0.475192\,Wh) + 0.52224\,Wh\\
\intertext{With our boundary estimation for $1 \le k \le 3$, this yields:}
min(E_{action}) &= 1 \cdot 0.475192\,Wh + 0.52224\,Wh = \underline{\underline{0.997432}}\,Wh\label{actionMA_min}\\
max(E_{action}) &= 3 \cdot 0.475192\,Wh + 0.52224\,Wh = \underline{\underline{1.947816}}\,Wh\label{actionMA_max}\\
%action_{MA_{min}} &= 1 \cdot 118798t \cdot 0.000004\frac{Wh}{token} + 10 \cdot 512t \cdot 0.000102\frac{Wh}{token}\\
%&= 0.475192Wh + 0.52224Wh\\
%&= \underline{\underline{0.997432}}Wh\label{actionMA_min}\\
%action_{MA_{max}} &= 3\cdot 118798t \cdot 0.000004\frac{Wh}{token} + 10 \cdot 512t \cdot 0.000102\frac{Wh}{token}\\
%&= 1.425576Wh + 0.52224Wh\\
%&= \underline{\underline{1.947816}}Wh\label{actionMA_max}
\end{align}

%We estimate a lower and an upper bound for the energy cost per action, where we estimate the the total number of tokens processed by DeBERTa to be equivalent to the number of tokens contained within the average HTML page ($k_{min}=1$) for the lower bound and three times the number of tokens within the average HTML page ($k_{max}=3$) for the upper bound, as detailed in the previous paragraph. This results in the following calculation for the energy cost $E_{action}$ for predicting a single action using MindAct.

\subsection{LASER}
In contrast to MindAct, LASER~\cite{ma2023laser} makes use of a proprietary language model, specifically GPT-4~\cite{openai2024gpt4technicalreport}. Since OpenAI has not published any numbers that would allow us to do precise estimations we have to rely on the best estimates made by external observers. LASER itself introduces states and state transitions for web agents, allowing to better recover from mistakes and restricting the possible \textit{actions} depending on the state of the agent. LASER uses one-shot prompting and makes the model think step-by-step to improve the models capabilities when dealing with complex user queries. Additionally, LASER has access to a memory buffer, to store and access intermediary results (previous \textit{actions}). Finally, it is specified that LASER is forced to produce a result after a maximum of 15 actions were generated.

\begin{figure}[ht]
    \centering
    \includegraphics[width=0.5\linewidth]{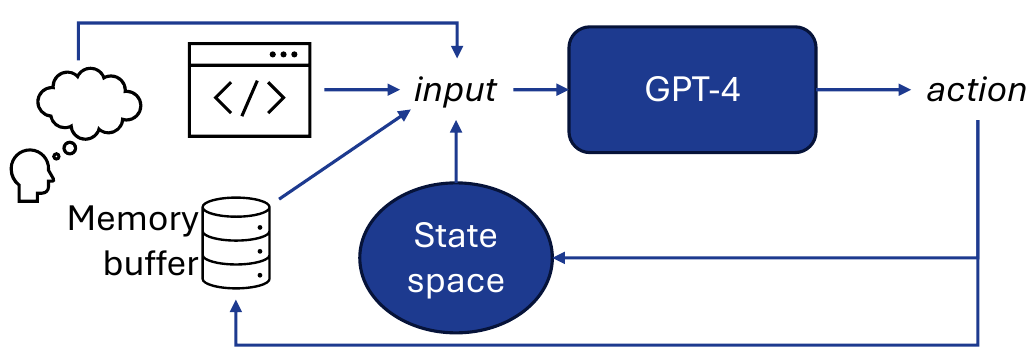}
    \caption{Pipeline depicting how an action is chosen in LASER. An \textit{input} consists of the user query, the HTML and a subset of actions the model can take depending on the state space. The LLM also has the option to access the memory buffer to aid recovery from failed paths. Memory buffer and state space get updated with the generated \textit{action}.}
    \label{fig:laser_pipe}
\end{figure}

However, the authors do not specify explicitly what the \textit{input} of their web agent is. We inferred, that they use the raw unmodified HTML by analyzing their results and WebShop~\cite{yao2022webshop}, the benchmark on which they tested their agent. %We come to this conclusion since the WebShop benchmark has two modes, HTML and simplified and the simplified mode is specifically mentioned for their test with ReAct~\cite{ma2023laser}.

To compare LASER to MindAct, we calculated the average number of tokens within a HTML page for the Mind2Web benchmark using the GPT-4~\cite{openai2024gpt4technicalreport} tokenizer at $\overline{N}_{GPT-4} = 93778$ tokens.
We estimate the energy per token $e_{GPT-4}$ through using reported token costs $C_{token}=10\,\$\frac{1}{10^6}$~\cite{openaipricing} and average energy costs in the US at $C_{energy}=0.16\,\$\frac{1}{kWh}$~\cite{electricity_prices} as a proxy. Note that we also assume that the energy costs only make about $k=50\%$ of the token pricing~\cite{model_cost}.

\begin{align}
    E_{action} &= \overline{N}_{GPT-4} \cdot e_{GPT-4} \label{firstlinelaser}\\
         \intertext{Where: } e_{GPT-4} &= k \cdot \frac{C_{token}}{C_{energy}}\\
    e_{GPT-4} &= 0.5 \cdot \frac{10\cdot 10^{-6}\,\$}{\frac{0.16\,\$}{kWh}}\\
    &= 0.03125\,Wh\\
    \intertext{Entering this into \ref{firstlinelaser}:}
    E_{action} &= 93778 \cdot 0.03125\,Wh\\
    %action_{LASER} &= 93778token \cdot k \cdot 0.0625\frac{Wh}{token}\\
    %&= k \cdot 5861.125Wh\\
    %&= 50\% \cdot 5861.125Wh\\
    &= \underline{\underline{2930.5625}}Wh\label{actionLASER_max}
\end{align}

\subsection{Comparing MindAct and LASER}
\label{sec:comparison}
Due to the lack of relevant information about the energy consumption of LLMs and the challenges involved in gathering this information as a third party, our results should be seen as estimations within a given range. 
For MindAct, we estimated upper and lower bounds (see \Cref{actionMA_max,actionMA_min}), where the upper bound consumes about two times the amount of energy compared to the lower bound. Compared to these, LASER consumes approximately 2900 times (lower bound of MindAct) and 1500 times (upper bound of MindAct) more energy.

When comparing the energy consumption per token for LASER and MindAct, it becomes clear that the preprocessing done in MindAct is vital. The energy consumption of Laser is only 600 times higher compared to flan-T5$_{XL}$ on its own. By utilizing the small and therefore energy efficient DeBERTa (which consumes 15000 times less energy per token than GPT-4) for preprocessing tasks, MindAct becomes significantly more energy efficient.

\subsection{Carbon Dioxide Emissions}
%'Find a mini van at Brooklyn City from April 5th to April 8th for a 22 year old renter.', 11 actions
%Find monthly daytime only parking nearest to Madison Square Garden starting from April 22, 7 actions
In the previous sections, we estimated the energy consumption per action. However, generally the tasks given to web agents require multiple actions to complete. In the Mind2Web benchmark the average number of actions per task is 7.3~\cite{deng2024mind2web}. This is a task taken from the training dataset of Mind2Web~\cite{deng2024mind2web} requiring 7 actions:
\begin{quote}
    \textit{Find monthly daytime only parking nearest to Madison Square Garden starting from April 22}
\end{quote}
Using the energy consumption estimated above, paired with the information on number of actions needed to complete a task, we can calculate the CO$_2$ emissions per task. In our calculations $\overline{N}_{actions}$ denotes the amount of actions taken to finish a task, $\text{\textit{Ø}}_{CO_2}$ the average CO$_2$ emission per Wh and is given as $\text{\textit{Ø}}_{CO_2}=0.453\,\frac{g}{Wh}$ based on the energy mix of the US~\cite{co2us}. Finally, $E_x$ denotes the energy consumed by web agent $x$ to take one action.
\begin{align}
Emission_{x} &= \overline{N}_{actions} \cdot \text{\textit{Ø}}_{CO2} \cdot E_{x}\\
&= 7.3 \cdot 0.453\,\frac{g}{Wh} \cdot E_{x}
\intertext{Lower bound for MindAct (\Cref{actionMA_min}):}
Emission_{min(MindAct)} &= 7.3 \cdot 0.453\,\frac{g}{Wh} \cdot 0.997432\,Wh = \underline{\underline{3.30}}\,g
\intertext{Upper bound for MindAct (\Cref{actionMA_max}):}
Emission_{max(MindAct)} &= 7.3 \cdot 0.453\,\frac{g}{Wh} \cdot 1.947816\,Wh = \underline{\underline{6.44}}\,g
\intertext{For LASER (\Cref{actionLASER_max}):}
Emission_{LASER} &= 7.3 \cdot 0.453\,\frac{g}{Wh} \cdot 2930.5625\,Wh = \underline{\underline{9691.08}}\,g
\end{align}

These values are equivalent to driving approximately between 13\,m and 25\,m for MindAct but a much larger distance of 39\,km for LASER with an average car, assuming 248.55\,g of CO$_2$ emissions per kilometer driven\footnote{\url{https://www.epa.gov/greenvehicles/greenhouse-gas-emissions-typical-passenger-vehicle\#typical-passenger}}.%https://www.umsteigern.de/wie-viel-ist-1-kilo-co2.html 211\,g of CO$_2$ per km

\section{A Plea for Dedicated Metrics and Small Models}
Our results in \Cref{sec:evaluation} highlight that the differences in energy consumption between models can be several orders of magnitude even for conservative estimations. While web agents are not widely used yet, they are bound to become a normal part of interacting with the internet in the future. To make their use sustainable at scale, we need to evaluate web agents with respect to their energy efficiency as well, enforcing a more holistic evaluation on the available benchmarks. To do to this, a standardized way of reporting energy consumption is necessary:

\begin{mybox}
\textbf{Dedicated Metrics:} For each LLM in the web agent's pipeline, provide number of tokens per action as well as the energy cost per token.
\end{mybox}

While these metrics will not allow for a perfectly accurate energy calculation, they serve as a valuable indicator for the web agents' sustainability. In addition, reporting these metrics aids in providing transparency about the energy consumption of web agents. Especially for agents powered by open source LLMs, it is feasible to readily run tests to obtain these numbers on local hardware.

Furthermore, reporting the energy consumption instead of the CO$_2$ emissions is advantageous, as it is independent of the respective energy mix of the country in which the evaluation is conducted. This further aids comparability while being easily convertible into more tangible values like CO$_2$ emissions using the respective energy mix for the calculation.

As shown in \Cref{sec:comparison}, smaller LLMs generally have a smaller energy footprint than larger ones.  While larger models are often more capable, the small models can perform well in predefined tasks, such as candidate generation in MindAct~\cite{deng2024mind2web}. Hence, using a two step approach where the small language model is employed to reduce the amount of tokens that a larger model has to process is beneficial for the web agents energy consumption. This approach also promotes a modular design of the web agent's pipeline, allowing to exchange parts individually, further improving sustainability and reproducibility. 

\begin{mybox}
\textbf{Ensemble of Small Models:} Use small language models when possible in the web agent pipeline to allow efficient preprocessing and an efficient modular structure.
\end{mybox}

\section{Limitations}
The accuracy and certainty of our calculations is limited by the availability of trustworthy information regarding the energy cost of the used LLMs. This is an issue especially for proprietary models, where it is not possible to run them locally and track the amount of energy used. While we can assume, that companies generally try to make these models cheaper to use to increase their profit margins, having them disclose this information would significantly increase the accuracy of such estimations. As such, we limited our calculation to a coarse and conservative estimation in terms of orders of magnitude. Doing so, allows us to still draw implications and provide generalizable recommendation for future web agents and their evaluation.

\section{Conclusion}
In this work we provide a comparison between energy consumption of two web agents, MindAct and LASER. We show that web agent design can influence the energy consumption by orders of magnitude and propose the introduction of standardized metrics to evaluate the energy consumption of web agents.
Our work has revealed several current gaps in web agent research, such as a lack of awareness for sustainability and transparency. We believe that accounting for sustainability in web agent research is important due to the environmental impact of these agents at scale. By using smaller models where possible and developing efficient algorithms, it is possible to reduce the amount of data the larger LLMs have to process, making the pipeline more energy efficient. In addition, we advocate for dedicated metrics to make the energy consumption of different web agents for the same task comparable. 
% \begin{acks}
% This research was supported by the AI4Europe project under grant number:
% This work is supported by the European Union’s Horizon Europe research and innovation program (HORIZON-CL4-2021-HUMAN-01) through the "SustainML" project (grant agreement No 101070408).
% \end{acks}

\bibliographystyle{ACM-Reference-Format}
\bibliography{bibliography}


\end{document}